\documentclass[11pt]{article}

\usepackage[margin=1in]{geometry}
\usepackage{needspace}
\usepackage{url}
\usepackage{amsmath}
\usepackage{amssymb}
\usepackage{graphicx}
\usepackage{algorithm}
\usepackage{algpseudocode}
\usepackage{booktabs}
\usepackage{xcolor}
\usepackage{wrapfig}
\usepackage[numbers,sort&compress]{natbib}
\usepackage[hidelinks]{hyperref}

\title{Multi-Rigid-Body Approximation of Human Hands with Application to Digital Twin}

\author{
Bin Zhao\textsuperscript{1}, Yiwen Lu\textsuperscript{1}, Haohua Zhu\textsuperscript{1}, Xiao Li\textsuperscript{2}, Sheng Yi\textsuperscript{1} \\[0.5em]
\textsuperscript{1}DexRobot Co. Ltd. \\
\texttt{ys@dex-robot.com} \\[0.5em]
\textsuperscript{2}Changchun Veterinary Research Institute, Chinese Academy of Agricultural Sciences, \\
State Key Laboratory of Pathogen and Biosecurity, \\
Key Laboratory of Jilin Province for Zoonosis Prevention and Control, Changchun, 130122, China \\
\texttt{skylee6226@163.com}
}

\date{}

\begin{document}

\maketitle

\begin{abstract}
Human hand simulation plays a critical role in digital twin applications, requiring models that balance anatomical fidelity with computational efficiency. We present a complete pipeline for constructing multi-rigid-body approximations of human hands that preserve realistic appearance while enabling real-time physics simulation. Starting from optical motion capture of a specific human hand, we construct a personalized MANO (Multi-Abstracted hand model with Neural Operations) model and convert it to a URDF (Unified Robot Description Format) representation with anatomically consistent joint axes. The key technical challenge is projecting MANO's unconstrained SO(3) joint rotations onto the kinematically constrained joints of the rigid-body model. We derive closed-form solutions for single degree-of-freedom joints and introduce a Baker-Campbell-Hausdorff (BCH)-corrected iterative method for two degree-of-freedom joints that properly handles the non-commutativity of rotations. We validate our approach through digital twin experiments where reinforcement learning policies control the multi-rigid-body hand to replay captured human demonstrations. Quantitative evaluation shows sub-centimeter reconstruction error and successful grasp execution across diverse manipulation tasks.

\end{abstract}

\section{Introduction}

Digital twin technology requires accurate real-time simulation of human manipulation, motivating hand models balancing visual realism with computational efficiency. Current approaches face a fundamental trade-off. High-fidelity mesh models like MANO \cite{romero2017embodied} and its extensions \cite{xie2024ms,ye2024physhand} provide excellent visual quality but require expensive soft-body simulation unsuitable for real-time applications. While optimization-based methods require 3.44ms, our approach achieves 5.36° mean error in 0.41ms—over 8$\times$ faster (Table~\ref{tab:projection_accuracy}). Skeleton-only approaches enable fast simulation but lack visual fidelity. Recent digital twin systems \cite{chen2024dexsim2real2,chen2024bidexhd} demonstrate strong task performance yet remain limited by hand models that cannot simultaneously maintain visual realism and high-frequency updates. This gap becomes critical in virtual training, teleoperation, and human-robot collaboration where both speed and visual accuracy matter.

We propose a multi-rigid-body approximation representing the hand as a kinematic tree of rigid links with fixed meshes, preserving MANO's (Multi-Abstracted hand model with Neural Operations) visual appearance while enabling standard rigid-body physics simulation. This maintains benefits of both approaches: the computational efficiency of rigid-body dynamics and visual quality of mesh-based models. The rigid links naturally map to URDF (Unified Robot Description Format) format, compatible with standard robotics simulators and control algorithms.

The technical challenge lies in bridging two different rotation representations. MANO parameterizes each joint as an unconstrained 3-DOF rotation in SO(3), while robotic joints are typically constrained to 1-DOF (hinge) or 2-DOF (universal) rotations. Simply discarding degrees of freedom loses essential motion, while naive projection onto constrained axes produces kinematically inconsistent results due to rotation non-commutativity. Previous work either accepts these approximation errors or relies on optimization-based retargeting that lacks real-time guarantees.

Our approach addresses this through a mathematically principled projection framework. For single-DOF joints, we derive a closed-form projection formula based on the tangent space structure of SO(3). For two-DOF joints, we develop an iterative method using the Baker-Campbell-Hausdorff (BCH) formula to handle the non-linear interaction between rotation axes. The method converges rapidly (typically 3-5 iterations) and produces kinematically consistent joint angles that best approximate the original MANO pose.

Our contributions are threefold. First, we present a complete pipeline from human hand capture to multi-rigid-body URDF model, including automated mesh segmentation and joint axis determination based on anatomical heuristics. Second, we develop closed-form and BCH-corrected projection methods for mapping unconstrained SO(3) rotations to kinematically constrained joints, with mathematical analysis of convergence and accuracy. Third, we validate our approach through digital twin experiments showing successful replay of human demonstrations using RL-trained policies, achieving sub-centimeter tracking error across diverse manipulation tasks.

\section{Related Work}

\noindent\textbf{Hand Modeling Approaches.} MANO \cite{romero2017embodied} established parametric hand representation by mapping pose and shape parameters to 3D meshes via linear blend skinning. Extensions include MS-MANO \cite{xie2024ms} with biomechanical constraints, MeMaHand \cite{wang2023memahand} combining parametric and non-parametric accuracy for two-hand reconstruction, and SMPL-X \cite{pavlakos2019expressive} for whole-body modeling with 54 hand joints. PhysHand \cite{ye2024physhand} uses multi-layer geometry with constraint-based dynamics for realistic contact. However, mesh-based methods require expensive soft-body simulation unsuitable for real-time use. Traditional rigid-body models enable faster simulation but lack visual fidelity. Our approach preserves MANO's visual quality while enabling rigid-body physics at 1000+ Hz.

\noindent\textbf{Motion Retargeting.} Converting between kinematic representations is challenging when projecting unconstrained rotations to constrained joints. Optimization methods \cite{ayusawa2017retargeting} achieve high accuracy but lack real-time guarantees. Analytical approaches \cite{darvish2019wholeretargeting} compute faster but often target specific morphologies. Rotation non-commutativity creates fundamental issues: naive projection yields kinematically inconsistent results, while the Baker-Campbell-Hausdorff formula \cite{hall2015lie} handles rotation composition but sees limited hand retargeting use. Neural methods like GeoRT \cite{yin2025geort} provide ultrafast retargeting but lack interpretability. We derive closed-form solutions for 1-DOF joints and BCH-corrected iteration for 2-DOF joints, achieving mathematical rigor and efficiency.

\noindent\textbf{Digital Twin and Simulation.} Digital twins require real-time hand simulation for teleoperation and collaboration. DexSim2Real\textsuperscript{2} \cite{chen2024dexsim2real2} constructs world models through active interactions, while BiDexHD \cite{chen2024bidexhd} achieves 74\% task success via multi-task RL. Isaac Gym reaches 30,000+ FPS for manipulation training versus MuJoCo's CPU limits, though MJX \cite{todorov2012mujoco} promises GPU acceleration. RL frameworks show strong results: DexMimicGen \cite{jiang2024dexmimicgen} generates 21K demonstrations from 60 human demos, ArtiGrasp \cite{zhang2024artigrasp} achieves 90\% simulation success with 67\% real-world transfer. These systems need hand models balancing visual realism with efficiency---the gap our multi-rigid-body approximation addresses.

\section{Method}

\subsection{Pipeline Overview}

\begin{figure}[t]
\centering
\includegraphics[width=\textwidth]{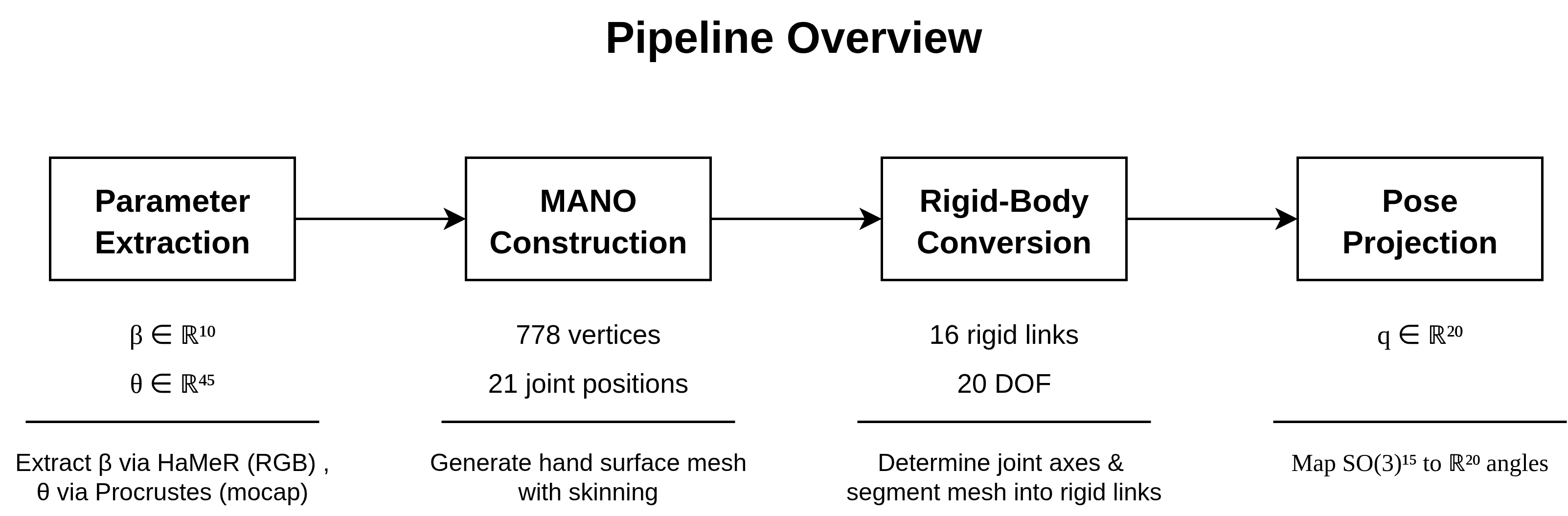}
\caption{Complete pipeline from human hand capture to multi-rigid-body URDF model. The pipeline processes optical motion capture or RGB input through four main stages: hand capture for recording human hand motion, MANO construction for fitting personalized models, rigid-body conversion for mesh segmentation and joint axis determination, and pose projection for mapping MANO poses (SO(3)$^{15}$) to URDF joint angles ($\mathbb{R}^{20}$).}
\label{fig:pipeline}
\end{figure}

Our pipeline processes human hand motion through four stages. First, we record motion using optical capture or RGB methods for high-fidelity kinematic data. Second, we fit a personalized MANO model to the captured hand shape, establishing correspondence between observed motion and parametric representation. Third, we segment the MANO mesh into rigid links and determine anatomically consistent joint axes for the multi-rigid-body approximation. Finally, we map MANO poses from the unconstrained SO(3)$^{15}$ space to the kinematically constrained URDF joint angles in $\mathbb{R}^{20}$.

\subsection{MANO Parameter Extraction}

The MANO model represents the hand as:
\begin{equation}
M(\beta, \theta) = W(T(\beta, \theta), J(\beta), \mathcal{W})
\end{equation}
where $\beta \in \mathbb{R}^{10}$ are subject-specific shape parameters extracted from RGB images, $\theta \in \mathbb{R}^{45}$ are pose parameters extracted from motion capture, $T$ is the shaped and posed mesh, $J$ are joint locations, and $\mathcal{W}$ are skinning weights.

Subject-specific shape parameters are extracted using HaMeR (Hand Mesh Recovery) \cite{pavlakos2024reconstructing}, a transformer-based method that regresses MANO parameters from RGB images. Hand pose parameters are extracted from optical motion capture markers through Procrustes alignment to MANO joint locations, involving anatomical frame construction and coordinate transformation to match MANO's kinematic conventions.

\subsection{Multi-Rigid-Body Construction}

\subsubsection{Frame Convention and Axis Determination}

Converting MANO to a multi-rigid-body representation requires defining joint frames and rotation axes for each joint. A key insight is that MANO, with fixed shape parameters $\beta \in \mathbb{R}^{10}$, can be viewed as a mapping $f_\beta: \text{SO(3)}^{15} \to \mathbb{R}^{778 \times 3}$ from joint rotations to vertex positions. Our multi-rigid-body model approximates this mapping on a lower-dimensional submanifold where joints have kinematic constraints.

We adopt wrist-aligned frames for all joints, meaning that all joint coordinate frames share the same orientation as the wrist frame (identity rotation). This choice maintains consistency with MANO's representation and simplifies the mathematical formulation. In this convention, the determination of rotation axes becomes the central challenge, as these axes must be expressed in the global wrist frame rather than local joint frames.

For anatomical plausibility, we constrain joint motions to principal movement patterns observed in human hands. The metacarpophalangeal (MCP) and carpometacarpal (CMC) joints exhibit two degrees of freedom corresponding to abduction-adduction and flexion-extension. The interphalangeal joints (PIP, DIP, IP) possess single degrees of freedom for flexion-extension only. These anatomical constraints guide our axis determination procedure.

\subsubsection{Axis Computation for Four-Finger Chains}

For the four fingers (index, middle, ring, pinky), we compute rotation axes based on the geometric structure of the hand at rest pose. The finger direction vector provides a natural reference:
\begin{equation}
z_f = \text{normalize}(J_{\text{PIP}}^f - J_{\text{DIP}}^f)
\end{equation}

The abduction axis must be perpendicular to the finger direction and aligned with the natural spreading motion. We determine this through a cross product with a reference direction that varies by finger to account for the fan-like arrangement of the metacarpals:
\begin{align}
y_f &= \text{normalize}(z_f \times \text{ref}_f) \\
x_f &= \text{normalize}(y_f \times z_f)
\end{align}
The reference vector $\text{ref}_f$ is chosen based on anatomical observations: index uses world vertical $[0, 0, 1]$, middle uses ring-to-index MCP vector (capturing the transverse arch), ring uses its MCP to middle MCP, and pinky references the ring MCP.

\subsubsection{Thumb Axis Determination}

The thumb's unique saddle joint at the CMC requires special treatment. The CMC joint enables opposition, the key movement distinguishing human manipulation. We determine the CMC axes through the thumb's geometric relationship with the index finger:
\begin{align}
y_{\text{CMC}} &= \text{normalize}(J_{\text{MCP}}^{\text{thumb}} - J_{\text{MCP}}^{\text{index}}) \\
z_{\text{CMC}} &= \text{normalize}(y_{\text{CMC}} \times (J_{\text{MCP}}^{\text{thumb}} - J_{\text{IP}}^{\text{thumb}}))
\end{align}
For the thumb MCP and IP joints, we apply a constant tilt angle of 55 degrees (0.96 radians) to account for the thumb's oblique orientation relative to the palm plane. This anatomically motivated tilt ensures realistic thumb flexion patterns.

\needspace{4\baselineskip}
\subsubsection{Mesh Segmentation}

\begin{wrapfigure}{r}{0.42\textwidth}
\vspace{-0.5em}
\centering
\includegraphics[width=0.38\textwidth]{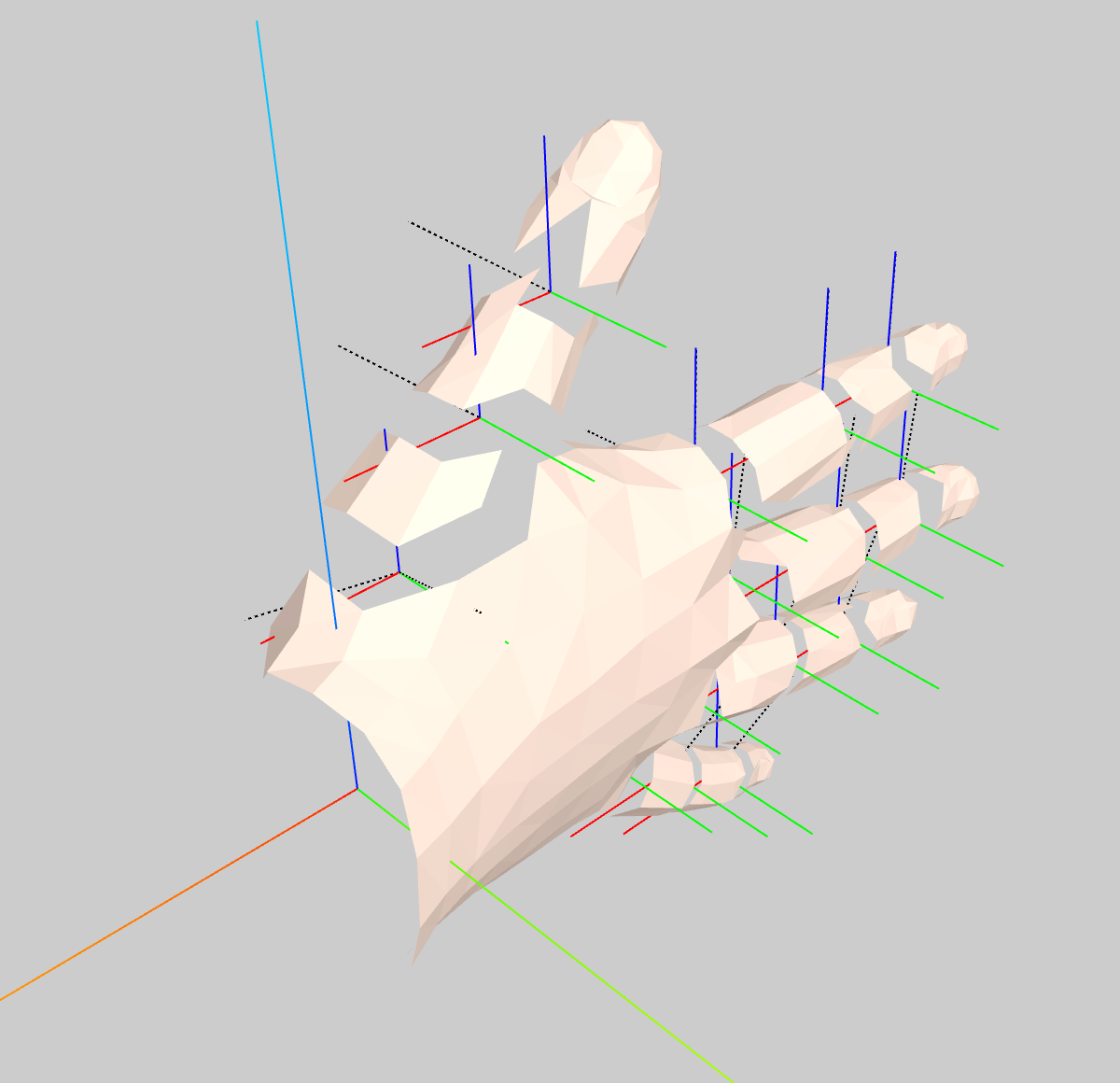}
\caption{Wrist-aligned coordinate system with anatomically determined rotation axes for finger joints. Color coding: Red=X axis, Green=Y axis, Blue=Z axis, black dashed lines indicate rotation axes.}
\label{fig:joint_frames}
\vspace{0.5em}
\end{wrapfigure}

With joint frames and axes defined, we segment the MANO mesh into rigid links. Each vertex $v$ with skinning weights $\mathcal{W}$ is assigned to its maximally weighted joint:
\begin{equation}
L(v) = \arg\max_{j} \mathcal{W}_{v,j}
\end{equation}
This produces 16 rigid segments (palm plus 15 phalanges), each with its local mesh geometry transformed to the corresponding joint origin for rigid attachment in the URDF model.

\subsection{Pose Projection: SO(3)$^{15}$ to $\mathbb{R}^{20}$}

Having defined the multi-rigid-body structure as an approximation of MANO, we now require a mapping from MANO's full SO(3)$^{15}$ pose space to the constrained $\mathbb{R}^{20}$ joint angle space of our URDF model. This mapping must preserve the essential motion characteristics while respecting the kinematic constraints. Our URDF model comprises five two-DOF joints at the MCP and CMC positions (contributing 10 parameters) and ten one-DOF joints at the interphalangeal positions (contributing 10 parameters).

We formulate this as a projection problem in SO(3). Given a rotation matrix $R \in \text{SO(3)}$ from MANO, we seek the joint angles that best approximate this rotation under the kinematic constraints. The projection must minimize the geodesic distance on the SO(3) manifold while remaining computationally efficient for real-time applications.

\subsubsection{Single-DOF Projection}

For joints constrained to rotate about a single axis $a \in \mathbb{S}^2$, we seek the angle $\theta$ that minimizes the Frobenius norm between the target rotation $R$ and the achievable rotation $R_a(\theta)$:
\begin{equation}
\theta^* = \arg\min_{\theta} \|R - R_a(\theta)\|_F^2
\end{equation}
Exploiting the differential structure of SO(3), we derive a closed-form solution. The key insight is that rotations around a fixed axis form a one-dimensional subgroup of SO(3), and the projection onto this subgroup can be computed analytically:
\begin{equation}
\theta = \text{atan2}(\langle \text{vee}((R - R^T)/2), a \rangle, (\text{tr}(R) - 1)/2)
\end{equation}
This formula leverages the decomposition of rotation matrices into symmetric and skew-symmetric parts. The skew-symmetric component $(R - R^T)/2$ encodes the instantaneous rotation axis and is proportional to $\sin\theta$, while the trace relates to the rotation angle through $\text{tr}(R) = 1 + 2\cos\theta$. The vee operator extracts the axial vector from the skew-symmetric matrix, and the inner product with $a$ projects onto the constrained axis.

\subsubsection{Two-DOF Projection with BCH Correction}

Joints with two degrees of freedom, such as the MCP and CMC joints, present a more complex challenge. Given axes $a_1$ and $a_2$ for abduction and flexion respectively, we seek angles $(\phi, \theta)$ satisfying:
\begin{equation}
R \approx R_{a_1}(\phi) \circ R_{a_2}(\theta) = \exp(\phi \cdot \hat{a}_1) \exp(\theta \cdot \hat{a}_2)
\end{equation}
where $\hat{a}_i$ denotes the skew-symmetric matrix corresponding to axis $a_i$.

The non-commutativity of rotations prevents simple independent projection. The Baker-Campbell-Hausdorff (BCH) formula quantifies this non-commutativity:
\begin{equation}
\log(\exp(X)\exp(Y)) = X + Y + \frac{1}{2}[X,Y] + \frac{1}{12}[X,[X,Y]] - \frac{1}{12}[Y,[X,Y]] + \cdots
\end{equation}
Truncating to first-order correction terms and recognizing that the Lie bracket in so(3) corresponds to the cross product of axial vectors $[\phi \cdot a_1, \theta \cdot a_2] = \phi\theta(a_1 \times a_2)$, we obtain an iterative refinement algorithm that accounts for the curved geometry of SO(3):

\begin{algorithm}
\caption{BCH-Corrected Two-DOF Projection}
\begin{algorithmic}[1]
\State $\omega \gets \log(R)$ \Comment{Convert to so(3)}
\State $\phi \gets \langle\omega, a_1\rangle$, $\theta \gets \langle\omega, a_2\rangle$ \Comment{Initialize}
\For{$i = 1$ to $3$} \Comment{Typically converges in 3 iterations}
    \State $\omega_{approx} \gets \phi \cdot a_1 + \theta \cdot a_2 + \frac{1}{2}\phi\theta(a_1 \times a_2)$
    \State $r \gets \omega - \omega_{approx}$ \Comment{Residual}
    \State $\phi \gets \phi + 0.5 \cdot \langle r, a_1\rangle$ \Comment{Update with relaxation}
    \State $\theta \gets \theta + 0.5 \cdot \langle r, a_2\rangle$
\EndFor
\State \Return $(\phi, \theta)$
\end{algorithmic}
\end{algorithm}

The cross-product term $\frac{1}{2}\phi\theta(a_1 \times a_2)$ captures the geometric effect of non-commutativity, ensuring kinematically consistent results.

\section{Experiments}

\subsection{Experimental Setup}

\subsubsection{Motion Capture System}

\begin{wrapfigure}[12]{r}[-1em]{0.35\textwidth}
\vspace{-1.5em}
\centering
\includegraphics[width=0.30\textwidth]{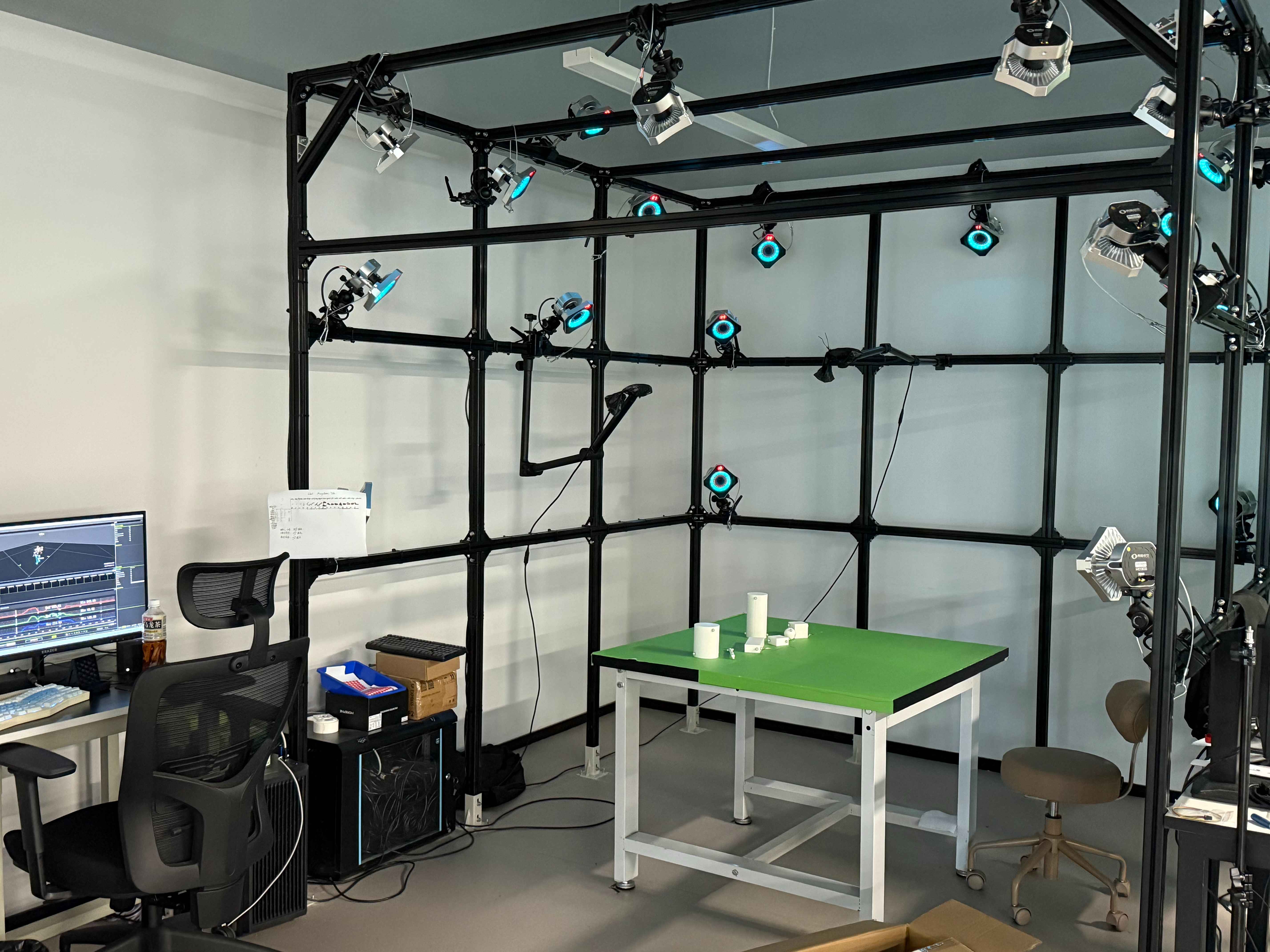}
\caption{Motion capture system configuration showing optical marker placement and capture volume.}
\label{fig:mocap}
\end{wrapfigure}

We capture hand manipulation using an optical motion capture system with 22 infrared cameras around a $2\text{m} \times 2\text{m}$ volume, sampling at 120 Hz with sub-millimeter accuracy. We attach 22 reflective markers to anatomical landmarks for precise joint trajectory reconstruction. Marker trajectories are fitted to MANO pose parameters using Procrustes alignment, achieving sub-centimeter accuracy.

\subsubsection{Digital Twin Methodology}

We validate through digital twin experiments where PPO policies in IsaacGym control the multi-rigid-body hand to reproduce human demonstrations. For each sequence: (1) convert mocap to MANO poses, (2) project to URDF angles, (3) train PPO to track trajectories, (4) evaluate accuracy and success. Actions are residual corrections accumulated via exponential filtering ($\tau = 0.9$). The reward balances tracking with regularization:
\begin{equation}
r_t = \exp(-\alpha_p \|p_o - \hat{p}_o\|) + \exp(-\alpha_r \cdot d_{\text{rot}}(R_o, \hat{R}_o)) - \|a_t\|^2
\end{equation}

\subsection{Results}

\subsubsection{Projection Accuracy}

We evaluate by projecting MANO poses to URDF angles and reconstructing via forward kinematics for comparison.

\begin{figure}[t]
\centering
\includegraphics[width=\textwidth]{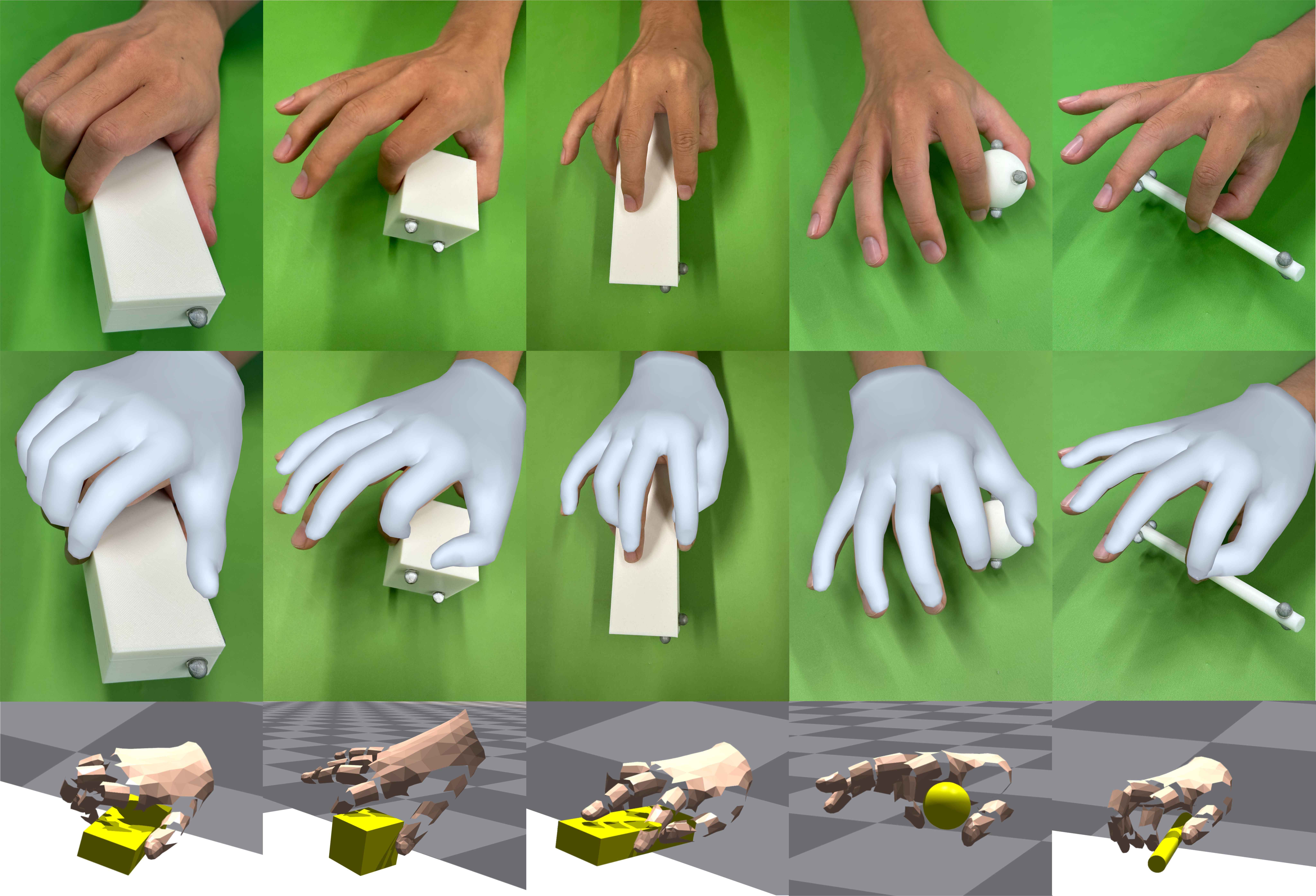}
\caption{Comprehensive evaluation results showing projection accuracy metrics, task success rates across different manipulation types, and visual comparison between human demonstrations and digital twin replay.}
\label{fig:results}
\end{figure}

\begin{table}[h]
\centering
\caption{Projection Accuracy Results}
\label{tab:projection_accuracy}
\begin{tabular}{@{}lcccc@{}}
\toprule
Method & Mean Error (°) & Max Error (°) & RMSE (°) & Time (ms) \\
\midrule
Our Method (BCH-corrected) & 5.36 & 21.46 & 24.31 & 0.41 \\
Naive Projection & 6.29 & 89.00 & 23.13 & 0.37 \\
Least Squares Projection & 5.32 & 21.46 & 24.31 & 3.44 \\
\bottomrule
\end{tabular}
\end{table}

Table~\ref{tab:projection_accuracy} shows round-trip errors for 100 poses. BCH-corrected achieves 5.36° mean error, nearly matching least-squares (5.32°) while 8.4$\times$ faster (0.41ms vs 3.44ms). Naive projection shows 18\% higher mean error (6.29°) and catastrophic failures (89° maximum), demonstrating the importance of handling rotation non-commutativity. BCH-corrected provides production-ready accuracy with real-time efficiency.

\subsubsection{Digital Twin Performance}

The rigid-body representation enables 1000+ Hz simulation (vs 10-30 Hz for soft-body MANO). We evaluate five manipulation primitives (Table~\ref{tab:grasp_success}), achieving 77.9\% success with 0.85cm tracking error.

\begin{table}[h]
\centering
\caption{Grasp Success Rates by Manipulation Type}
\label{tab:grasp_success}
\small
\begin{tabular}{@{}lcc@{}}
\toprule
Manipulation Type & Success Rate (\%) & Tracking Error (cm) \\
\midrule
Power Grasp & 73.0 & 0.83 \\
Precision Pinch & 82.3 & 0.95 \\
Lateral Pinch & 80.2 & 0.75 \\
Cylindrical Grasp & 72.2 & 0.87 \\
Spherical Grasp & 81.6 & 0.87 \\
\midrule
Overall Average & 77.9 & 0.85 \\
\bottomrule
\end{tabular}
\end{table}

\section{Conclusion}

We presented a complete pipeline for multi-rigid-body hand approximations balancing visual fidelity with computational efficiency. Our key contribution is a mathematically principled framework projecting unconstrained SO(3) rotations onto constrained joints, using closed-form solutions for 1-DOF joints and BCH-corrected iteration for 2-DOF joints. Experiments demonstrate accurate digital twin replay with sub-centimeter tracking error while maintaining real-time performance. The method bridges human motion capture and robot simulation, facilitating teleoperation, virtual training, and human-robot collaboration.

Future work includes extending to dynamic grasping forces, investigating learned projection methods adapting to individual hand kinematics, and applying the framework to retarget motions across different robot hand designs.

\section*{Acknowledgments}

This work will be published in the Proceedings of the 2025 International Conference on Biomechanical Systems and Robotics (ICBSR'25), Springer Lecture Notes in Mechanical Engineering (LNME).

\bibliographystyle{plain}
\bibliography{references}

\end{document}